\documentclass[aps,prx,twocolumn,
			   groupedaddress,superscriptaddress,
			   amsfonts,amssymb,amsmath,
			   citeautoscript,longbibliography,
			   a4paper,nofootinbib
			   ]{revtex4-1}
\usepackage[utf8]{inputenc}
\usepackage[english]{babel}

\usepackage{microtype} %For better kerning and symbol-stretching
\usepackage{xspace} %For the \xspace command

\usepackage{txfonts}  %Times Roman fonts
\usepackage{txfontsb} %Addition for txfonts, including old style numerals and greek

\usepackage{bm} %Bold math symbols with \bm{} (Greek and other symbols)

\usepackage{xcolor}
\usepackage[]{graphicx} % "demo" option to disable rendering for speed
\graphicspath{{figs/}}

\usepackage[]{booktabs}
\usepackage{array}
\usepackage{layouts}
\usepackage{multirow}

\usepackage{enumerate}
\usepackage[inline]{enumitem}

\usepackage{hyperref}
\hypersetup{colorlinks,
	linkcolor={blue!75!black!80!yellow},
	citecolor={blue!75!black!80!yellow},
	urlcolor={blue!75!black!80!yellow}
}

\usepackage[capitalize,nameinlink]{cleveref}

\crefname{subequations}{Eqs.}{Eqs.} %Specific changes to allow for Eqs.-wording when referring to a set of subequations. Label of subequations must include [subequations] as an option.
\Crefname{subequations}{Eqs.}{Eqs.}
\crefformat{subequations}{#2Eqs.~(#1)#3}
\Crefformat{subequations}{#2Eqs.~(#1)#3}
\crefname{page}{p.}{p.} %Changing from 'page' to 'p.'

\usepackage{placeins}

\usepackage{siunitx}
\DeclareSIUnit[number-unit-product = ]\percent{\char`\%} % remove spacing for \percent

\usepackage[centering,hmargin=16mm,tmargin=30mm,bmargin=26mm]{geometry}

\usepackage{soul}

\usepackage{textcomp} % for \textrightarrow
\usepackage{xifthen}
\usepackage{etoolbox}
\newboolean{togglecomments}
\newboolean{togglechanges} 

\setboolean{togglecomments}{true}  
\setboolean{togglechanges}{false} 

\newcommand{\textblacksquare}{$\blacksquare$}
\newcommand{\comment}[2]{\ifbool{togglecomments}%
		{\textcolor{blue!70!black}{\small\sf\textsuperscript{\textsc{\textsf{#1}}}[#2]}} % if true, show comments
		{}}     % if false, do nothing
\newcommand{\swap}[2]{\ifbool{togglechanges}
	{#2}  % TC-only version
	{\textcolor{red!70!black}{[#1]}\textrightarrow{}\textcolor{green!50!black}{[#2]}}}
\newcommand{\remove}[1]{\ifbool{togglechanges}
	{}    % TC-only version
	{\textcolor{red!70!black}{#1}}}
\newcommand{\inset}[1]{\ifbool{togglechanges}
	{#1}  % TC-only version
	{\textcolor{green!50!black}{#1}}}
\newcommand{\citeremind}[1]{%
	[\textcolor{red!75!black!80!yellow}{\textblacksquare%
		\ifthenelse{\isempty{#1}}{}{\textsuperscript{\tiny\textsf{#1}}}%
	}]\xspace}

\newcommand{\rv}{\mathbf{r}}

\newcommand{\e}{\mathrm{e}}

\newcommand{\ie}{i.e.,\@\xspace} %Gobble-spaces of the "small" type (small is ensured by adding \@)

\newcommand{\eg}{e.g.,\@\xspace}

\newcommand{\appropto}{\mathrel{\vcenter{
			\offinterlineskip\halign{\hfil$##$\cr
				\propto\cr\noalign{\kern.2pt}\sim\cr\noalign{\kern-2.5pt}}}}}

\makeatletter
\newcommand{\raisemath}[1]{\mathpalette{\raisem@th{#1}}}
\newcommand{\raisem@th}[3]{\raisebox{#1}{$#2#3$}}
\makeatother

\DeclareFontFamily{U}{mathx}{\hyphenchar\font45}
\DeclareFontShape{U}{mathx}{m}{n}{<5> <6> <7> <8> <9> <10>
                                  <10.95> <12> <14.4> <17.28> <20.74> <24.88>
                                  mathx10}{}
\DeclareSymbolFont{mathx}{U}{mathx}{m}{n}
\DeclareFontSubstitution{U}{mathx}{m}{n}
\DeclareMathAccent{\widebar}{0}{mathx}{"73}

\renewcommand{\paragraph}[1]{\vskip 1ex\noindent\textbf{#1.}~}

\usepackage[eulergreek]{sansmath}
\makeatletter
\renewcommand\@make@capt@title[2]{%
    \@ifx@empty\float@link{\@firstofone}{\expandafter\href\expandafter{\float@link}}%
    \sansmath\sffamily\textbf{#1\@caption@fignum@sep}#2 % does not work with the newtx* packages unfortunately
}%

\makeatother

\makeatletter
    \DeclareRobustCommand*{\deactivateaddvspace}{\let\addvspace\@gobble} % "deactivates" \addvspace command
    \DeclareRobustCommand*{\deactivatetocsubsections}{
    \def\l@subsection##1##2{}    % these definitions are inherited from \l@@sections, see 
    \def\l@subsubsection##1##2{} % ltxutils.dtx; "reset" them to remove subsections in ToC
    }
\makeatother

\usepackage{lineno}
\begin{document}
%-----------------
%----- TITLE -----
%-----------------
\title{Multimodal Foundation Models for Material Property Prediction and Discovery}%\thanks{This article has been accepted for publication in *Newton* (Cell Press). The final version is available at \href{https://doi.org/10.1016/j.newton.2025.100016}{DOI: 10.1016/j.newton.2025.100016}.}

\onecolumngrid
\begin{center}
    \textbf{\large This article has been accepted for publication in \textit{Newton} (Cell Press)}\\
    The final version is available at \href{https://doi.org/10.1016/j.newton.2025.100016}{DOI: 10.1016/j.newton.2025.100016}.
\end{center}
\vspace{1em}
\twocolumngrid

%------------------------------------
%----- AUTHORS AND AFFILIATIONS -----
%------------------------------------
\def\mitaffil{Department of Physics, Massachusetts Institute of Technology, USA}
\def\miteecsaffil{Department of Electrical Engineering and Computer Science, Massachusetts Institute of Technology, USA}
\def\dtuaffil{Department of Electrical and Photonics Engineering, Technical University of Denmark, Denmark}
\def\ucaffil{Data Science Institute, University of Chicago, USA}
\def\jhapl{Research and Exploratory Development, John Hopkins University Applied Physics Laboratory, USA}
\def\equalcontrib{These authors contributed equally to this work.}
\def\corres{\{vmoro, soljacic\}@mit.edu}

\author{Viggo~Moro~$^{*, \dag}$}
\affiliation{\mitaffil}
\author{Charlotte~Loh~$^*$}
% \thanks{\corres}
\affiliation{\miteecsaffil}
\author{Rumen~Dangovski~$^*$}
\affiliation{\miteecsaffil}
\author{Ali~Ghorashi}
\affiliation{\mitaffil}
\author{Andrew~Ma}
\affiliation{\miteecsaffil}
\author{Zhuo~Chen}
\affiliation{\mitaffil}
\author{Samuel~Kim}
\affiliation{\jhapl}
\author{Peter~Y.~Lu}
\affiliation{\ucaffil}
\author{Thomas~Christensen}
\affiliation{\dtuaffil}
\author{Marin~Solja\v{c}i\'{c}~$^\dag$}
% \thanks{\corres}
\affiliation{\mitaffil}
%---------------------------
%----- KEYWORDS & PACS -----
%---------------------------
\keywords{}
\pacs{}

%--------------------
%----- ABSTRACT -----
%--------------------
\begin{abstract}
Artificial intelligence is transforming computational materials science, improving the prediction of material properties, and accelerating the discovery of novel materials. 
Recently, publicly available material data repositories have grown rapidly.
This growth encompasses not only more materials but also a greater variety and quantity of their associated properties.
Existing machine learning efforts in materials science focus primarily on single-modality tasks, \ie relationships between materials and a single physical property, thus not taking advantage of the rich and multimodal set of material properties.
Here, we introduce {Multimodal Learning for Materials} (MultiMat), which enables self-supervised multi-modality training of foundation models for
materials.
We demonstrate our framework's potential using data from the Materials Project database on multiple axes:
\begin{enumerate*}[label=(\roman*)]
    \item MultiMat achieves state-of-the-art performance for challenging material property prediction tasks; 
    \item MultiMat enables novel and accurate material discovery via latent space similarity, enabling screening for stable materials with desired properties; and 
    \item MultiMat encodes interpretable emergent features that may provide novel scientific insights.
\end{enumerate*}
\end{abstract}

\maketitle
\def\thefootnote{*}\footnotetext{\equalcontrib}
\def\thefootnote{\dag}\footnotetext{\corres}

%------------------------
%----- INTRODUCTION -----
%------------------------

\section{Introduction}

\begin{figure*}[!tb]
	\centering
	\includegraphics[scale=1.0]{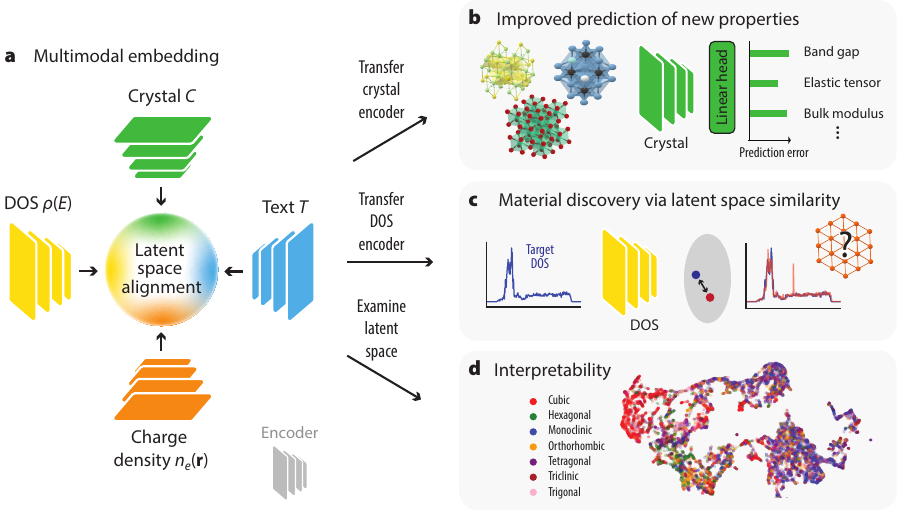}
	\caption{%
	    \textbf{The Multimodal Learning for Materials (MultiMat) approach.}
     \textbf{a},~Crystal ($C$), DOS ($\rho(E)$), charge density ($n_e(\mathbf{r})$), and text ($T$) encoders map each modality to embeddings in a shared multimodal latent space (center).
     MultiMat's training objective aligns the embeddings of different modalities corresponding to the same material.
     \textbf{b},~Application of MultiMat in improved prediction of materials' properties. The $C$ encoder from (a) is transferred, and a randomly initialized linear head is trained jointly with the transferred encoder to predict material properties.
     \textbf{c},~Application of MultiMat in material discovery. The DOS encoder embeds a target DOS (in blue). In the shared latent space, the closest crystal embedding (in red) from a large collection of crystal embeddings is selected. Since the embeddings of DOS and crystal are aligned during training, the crystal whose embedding is closest to the target DOS embedding is highly likely to have a DOS (in red) that closely resembles the target. Therefore, this crystal is identified as the best candidate. \textbf{d},~Application of MultiMat in enabling interpretability. We visualize the latent space of the crystal encoder using dimensionality reduction to reveal information about properties of materials that are implicitly encoded in the embeddings.
	}
    \label{fig:method}
\end{figure*}

Data-based approaches have become increasingly prevalent in computational materials science~\cite{ghiringhelli2015big,Ward_2016,sun2019map,deringer2021origins,zhong2020accelerated,butler2018machine,damewood2023representations}, due to the rapid algorithmic innovations in the field of machine learning (ML)~\cite{goodfellow2016deep} as well as by the growing amount of data available in materials science databases~\cite{hellenbrandt2004inorganic,10.1063/1.4812323, article_snumat, tang2019comprehensive, zhang2019catalogue, vergniory2019complete}.
An exciting aspect of ML in materials science lies in its potential to greatly accelerate calculations. Although training an ML model requires an up-front computational cost, predicting a material property using a trained ML model is substantially faster than running an ab initio calculation~\cite{schleder2019dft,axelrod2022learning,huang2023central}.
%This advantage in computational speed is crucial for 
The discovery of new materials relies on that speedup since the vast combinatorial space of possible materials makes exhaustive ab initio calculations computationally infeasible.
There have been a number of works that demonstrate the use of ML models to rapidly screen large amounts of materials with the aim of accelerating materials discovery~\cite{saal2020machine,gomez2016design,lu2018accelerated,ma2023topogivity}.
Beyond these screening-based approaches---which rely on predictive models---there is also an emerging interest in the use of generative models for materials discovery~\cite{fuhr2022deep,anstine2023generative,yao2021inverse}. 
Developing better graph neural networks (GNNs)~\cite{Xie_2018,schutt2018schnet,doi:10.1021/acs.chemmater.9b01294,Choudhary_2021,yan2022periodic,lin2023efficient} has represented the research frontier for achieving state-of-the-art predictive performance of materials. However, while interpretability has been a focus of ML for science, including in the domain of materials~\citep{ma2023topogivity,oviedo2022interpretable,allen2022machine,wang2022crabnet,hargreaves2020earth,zhong2022explainable,muckley2023interpretable}, GNNs, as any other deep neural network, usually fall short when it comes to interpretability.

An increasingly important paradigm in ML is foundation models, which are general-purpose ML models that are pre-trained on large amounts of data and then fine-tuned for a variety of applications~\citep{bommasani2021opportunities}.
Notable examples include GPT-4~\cite{gpt4} and Gemini~\citep{team2023gemini}.
Because pre-training is performed using unsupervised methods, these foundation models are able to take advantage of extremely large amounts of data that would normally be difficult to utilize when directly training models for specific downstream tasks.
A seminal work in multimodal learning is Contrastive Language Image Pre-training (CLIP)~\cite{radford2021learning}, which can be used to train multimodal foundation models.
CLIP aligns an image encoder with a text encoder, encouraging the embeddings of the image and captions to be similar.
Subsequent efforts~\cite{zhai2023sigmoid, li2022blip, zhong2021regionclip, gao2021clipadapter, ramesh2022hierarchical}, %work has 
have predominantly focused on multimodal learning with just two modalities (usually images and text)~\cite{kim2021vilt, wang2022simvlm, yu2022coca, yuan2021florence}.
How to best incorporate more than two modalities remains an open problem~\cite{girdhar2023imagebind, xue2023ulip, guzhov2021audioclip}.

Here, we adapt CLIP to the materials domain and also extend it to multimodal pre-training with an arbitrary number of modalities.
We leverage the fact that materials databases are inherently multimodal: \eg besides the crystal structure, the density of states (DOS)~\cite{TORIYAMA2022100002, lee2023density} and charge density~\cite{DOSSANTOS2020127431} convey rich information about materials. 
Textual descriptions of the crystal, which can be machine-generated~\cite{rubungo2023llm}, offer a fourth modality that is additionally computationally cheap to acquire. It is important to point out that the above-mentioned material modalities are not information-independent, since they can be computed from the crystal structure. The same holds true for image-caption pairs that were used in CLIP. Therefore, the point of contrastive multimodal pre-training is not to leverage modalities with independent information but rather to learn better representations by integrating different perspectives of the same underlying data~\cite{pmlr-v119-wang20k, chen2020simple, oord2019representation, daunhawer2023identifiabilityresultsmultimodalcontrastive}. Motivated by these opportunities, we introduce \emph{Multimodal Learning for Materials} (MultiMat), a novel framework for training a foundation model for crystalline materials that allows for the incorporation of several modalities. 
The basis for MultiMat is a multimodal pre-training method that connects high-dimensional material properties (\ie modalities) in a shared latent space to produce highly effective material representations that can then be transferred to various downstream tasks. Using MultiMat, we pre-train a state-of-the-art GNN on the Materials Project~\cite{10.1063/1.4812323} database to demonstrate its ability to produce state-of-the-art foundation models for materials. Very recently, preliminary work has explored related ideas for molecules~\cite{takeda2023multi} and a structure-agnostic multi-task learning approach for crystals~\cite{prein2023mtencoder}.

The MultiMat framework trains a foundation model for materials by aligning the latent spaces of encoders of different information-rich modalities, such as the crystal structure, DOS, charge density, and textual description, as shown in \cref{fig:method}a. 
This alignment process produces shared latent spaces and effective material representations which can then be leveraged for a series of downstream tasks (\cref{fig:method}b--d).
For instance, the crystal encoder can be transferred and fine-tuned for material property prediction, enabling improved predictive performance compared to traditional training techniques.
Since MultiMat aligns the latent spaces of different modalities, it can also be used in a novel material discovery strategy by screening large crystal-structure databases with comparisons between target properties and candidate crystals based on the latent-space similarity. 
Finally, we demonstrate the interpretability enabled by the MultiMat approach, by exploring the latent space from MultiMat using a dimensionality reduction approach.

\section{Results}

\subsection{Modalities and Architecture}
To illustrate the MultiMat framework, we consider four modalities for each material, all from the Materials Project database
\begin{enumerate*}[label=(\roman*)]
\item the crystal structure, which we denote by $C=(\{(\rv_i, E_i)\}_i, \{\mathbf{R}_j\}_j)$, where $\{(\rv_i, E_i)\}_i$ is a set containing the coordinates $\rv_i$ and chemical element $E_i$ of the $i$-th atom in the unit cell, and $\{\mathbf{R}_j\}_j$ is the set of unit cell lattice vectors; 
\item the DOS, $\rho(E)$, as a function of energy $E$; 
\item the charge density $n_{e}(\rv)$ as a function of position $\rv$; and
\item a textual description $T$ of the crystal obtained from Robocrystallographer~\cite{robocrystallographer}.
\end{enumerate*}
For each material modality, we train a separate neural network encoder to learn a parameterized transformation from raw data to an embedding in a shared latent space.
The $C$ encoder uses PotNet, a state-of-the-art GNN~\citep{lin2023efficient}; 
the encoders of $\rho(E)$ and $n_{e}(\rv)$ are based on the Transformer~\cite{vaswani2023attention} and 3D-CNN architectures~\cite{xie2017aggregated}. The $T$ encoder uses a frozen MatBERT~\cite{matbert} model, a Bidirectional Encoder Representations from Transformers (BERT) textual model~\cite{devlin2019bert} that has been pre-trained on material science literature.
A key advantage of the $T$ modality is that its data collection is relatively low cost since Robocrystallographer can be used to generate a $T$ modality for every $C$; thus $T$ can be used to obtain a much larger pre-training dataset.
Conversely, $T$ may not contain as rich information as other ``high-cost'' modalities like $\rho(E)$ and $n_{e}(\rv)$, which are usually obtained from ab initio simulations.
Additional modality and architecture details are provided in the \hyperref[sec:methods]{Methods} section.

\subsection{Overview of Multimodal Pre-training Methods}
\label{sec:overview_methods}
MultiMat adapts CLIP~\cite{radford2021learning} to the materials science domain through several extensions that allow for the integration of more than two modalities.
Below, we give a brief summary of CLIP and these extensions (see \hyperref[sec:methods]{Methods} for additional details): 
\begin{description}[leftmargin=1.5\parindent]
    \item[\textbf{CLIP}]
    Applies to two modalities. We adapt CLIP to materials science by replacing the traditional image--text pairs with $C$ paired with one other modality in $\{\rho(E),n_{e}(\rv),T\}$. CLIP encourages alignment between the embeddings of a pair of modalities (via the loss term in \cref{eq:CLIP_single}; \hyperref[sec:methods]{Methods})
    \item[\textbf{AllPairsCLIP}]
    When there are more than two modalities involved, multiple pairs of modalities can be created. AllPairsCLIP includes the pairwise CLIP loss between \emph{all} possible pairwise combinations of modalities; the loss is averaged over all such pairs. 
    \item[\textbf{AnchoredCLIP}]
    Because AllPairsCLIP considers all possible pairwise combinations, the number of loss terms increases significantly with more modalities. 
    A cheaper alternative is to only average over pairwise combinations that include $C$, \ie the `anchor', since the $C$ encoder is arguably the most crucial for transferring to other downstream tasks (\eg prediction tasks typically use the crystal structure as inputs). 
\end{description}

The loss terms of both AllPairsCLIP and AnchoredCLIP are aggregates of pairwise loss terms; for $n$ modalities, they feature $n(n-1)/2$ and $n-1$ individual pairwise loss terms, respectively. In the Supplementary Information, we explore other methods that align three or more modalities without pairwise decomposition (\ie there is only a single loss term regardless of the number of modalities).

A central advantage of pairwise alignment is the ability to exploit all available modality pairs, even when some pairs may be missing for certain database entries (since these loss terms can simply be set to zero).
This is an important feature since the coverage of material databases is often incomplete: \eg some entries may only have information for $C$ and $\rho(E)$, others only for $C$ and $n_e(\rv)$.
A pairwise multimodal loss allows MultiMat to take advantage of a greater total amount of data than would be possible with non-pairwise methods since the information of incompletely covered entries can still be incorporated.

\subsection{Crystal Property Prediction}

\begin{figure}[htb]
	\centering
	\includegraphics[scale=1]{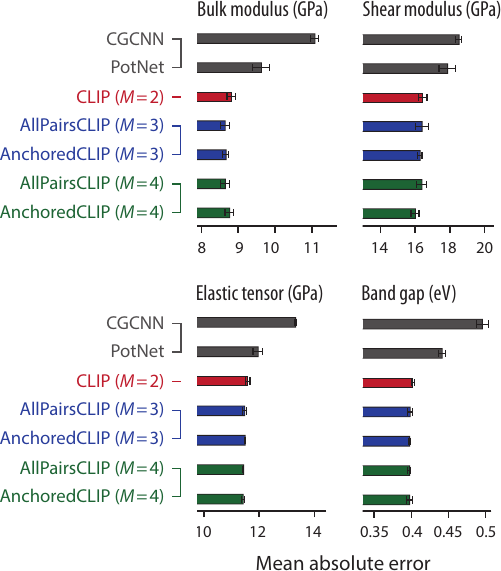}
	\caption{%
        \textbf{Crystal property prediction.} 
        Mean absolute error (MAE) for the prediction of various crystal properties across baseline methods and MultiMat.
        Methods are grouped by color according to the number of modalities, $M$, selected from the set of all modalities $\{ C, \rho(E), n_{e}(\rv), T\}$ (with $C$ always selected). Results for the $M=2$ and $M=3$ cases show the average performance over all allowed combinations for each category (individual experiments reported in the Supplementary Information) and error bars give the standard deviation over 3 random seeds, averaged over all experiments within that category. 
        } 
    \label{fig:property_pred}
\end{figure}

After the multimodal alignment stage in MultiMat, the $C$ encoder can be fine-tuned on various predictive tasks by attaching a randomly initialized linear head and fine-tuning end-to-end.
We explore the tasks of predicting the bulk modulus, shear modulus, elastic tensor, and band gap corresponding to a crystal input. 
The mechanical property tasks use the Materials Project database~\cite{10.1063/1.4812323} and the band gap task uses the SNUMAT semiconductor database~\cite{article_snumat}. These tasks were chosen because they have relatively few labeled data points compared to the number of data points used during pre-training. In particular, roughly \num{154000} data points are used during pre-training compared to roughly \num{7000} data points for the bulk modulus, shear modulus, and elastic tensor tasks and roughly \num{10000} data points for the band gap task.
Note that for the crystal property prediction tasks, only the crystal structure is used (and not any of the other modalities used for multimodal pre-training).

\Cref{fig:property_pred} compares MultiMat using multimodal pre-training with \numrange{2}{4} modalities against baselines without multimodal pre-training. 
The two baselines are CGCNN~\citep{Xie_2018}, the first method using GNNs for crystal property prediction, and PotNet~\citep{lin2023efficient}, the current state-of-the-art method for crystal property prediction using GNNs.
Note that MultiMat also uses the architecture of PotNet for the $C$ encoder. 
For the two- and three-modality cases in \cref{fig:property_pred}, the shown results are averages of experiments over all possible two- or three-modality combinations from the set $\{ C, \rho(E), n_{e}(\rv), T\}$ with $C$ always chosen.
For example, for two modalities, results are the average MAE over the combinations $\{ C,\rho(E) \}$, $\{ C,n_{e}(\rv) \}$, and $\{ C, T \}$ (results of individual experiments are shown in the Supplementary Information).
MultiMat pre-training significantly improves predictive performance compared to the baselines that do not make use of any pre-training. In particular, MultiMat reduces the MAE by up to ${\sim}\,10\%$ compared to PotNet, which is the current state-of-the-art and the $C$ architecture used in MultiMat.
This performance improvement is comparable to the improvement when going from CGCNN to PotNet, methods that are separated by five years that represent the first method and state-of-the-art method for crystal property prediction respectively. Moreover, note that MultiMat is a pre-training method that can be used on top of any existing or future crystal encoder to substantially improve its performance. Thus, the primary focus is on the performance difference between PotNet and MultiMat, with the CGCNN baseline serving to contextualize the overall performance advancements.

We observe that including three or more modalities during pre-training marginally improves performance over just two modalities (see~\cref{fig:property_pred}). 
On the other hand, we found no significant gains in using MultiMat with $M=4$ modalities over $M=3$ modalities. 
Speculatively, this might reflect that:
(i)~the fourth modality offers only a marginally different perspective on the material compared to the other three modalities or
(ii)~the current implementation is not able to take advantage of the additional modality due to model capacity limitations (to ensure fair comparison, we use a fixed architecture across all experiments and do not increase the neural network capacity to match the corresponding increased complexity).

\begin{figure}[htb]
	\centering
	\includegraphics[scale=1]{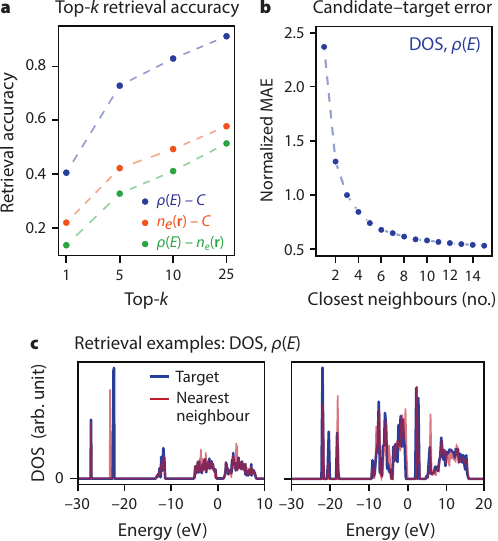}
	\caption{%
        \textbf{Material discovery via latent space similarity.} 
        \textbf{a},~Top-$k$ accuracies for cross-modality retrieval using encoders pre-trained with AnchoredCLIP, averaged over the test set.
        \textbf{b},~Normalized MAE between the target $\rho(E)$ from the test set and the $\rho(E)$ corresponding to the best crystal candidate from the training set, identified through our latent space similarity approach when the number of closest neighbors considered is varied. The best crystal candidate is selected from a set of crystals whose embeddings are the closest neighbors to the target $\rho(E)$ in the shared latent space, where the chosen crystal has a $\rho(E)$ with the smallest normalized MAE compared to the target $\rho(E)$. MAE values are normalized by the area of target $\rho(E)$ (both computed in the $(-5 \ \textrm{eV}, 5 \ \textrm{eV})$ range) and the values reported here are averaged over the whole test set.
        \textbf{c},~Two examples of the $\rho(E)$ corresponding to the best $C$ candidate found via latent space similarity overlaid with the target $\rho(E)$ of the material discovery process.
        } 
        \label{fig:inverse_design}
\end{figure}

\subsection{Material Discovery via Latent Space Similarity}

A key motivation for building fast surrogate predictive models is to enable accelerated design or identification of materials with specified properties.
In this section, we demonstrate an example of how MultiMat can achieve this goal via latent-space similarity, by screening a large material database and selecting the candidate which possesses the highest similarity to the desired property.  
Taking the example of identifying a material with a specific ``target'' DOS, we proceed by:
(1)~embedding the target DOS using the $\rho(E)$ encoder;
(2)~embedding each crystal in the database of candidate materials using the $C$ encoder; and
(3)~identifying the top-$k$ crystals that maximize the (cosine) similarity between the $\rho(E)$ and $C$ embeddings.
\Cref{fig:inverse_design} presents results for material discovery via latent space similarity for a MultiMat model trained using AnchoredCLIP with the three modalities $C$, $\rho(E)$, and $n_{e}(\rv)$.

We first investigate how well the latent spaces of the different encoders are aligned since good alignment is crucial for selecting good candidate materials. To this end, we explore the cross-modality retrieval performance of the model, \ie how often the model given a sample of a certain modality was able to retrieve the sample of another modality corresponding to the same material; the results are shown in \cref{fig:inverse_design}a. 
Retrieval performance was measured on a test set containing roughly \num{15000} materials (that all have $C$, $\rho(E)$, and $n_{e}(\rv)$ entries in the Materials Project database).
``DOS--crystal'' at top-$k$ refers to the average accuracy (over all DOS samples in the test set) that the correct crystal structure is present within the top-$k$ samples retrieved given a DOS sample in the test set.
The challenge of the retrieval task depends on the size of the dataset for which retrieval is performed (in our case consisting of roughly \num{15000} materials) and can be viewed as a classification task where the number of classes equals the number of samples in the dataset. Considering the size of our dataset, the strong retrieval performance demonstrates that MultiMat achieves effective alignment between the encoders of the different modalities.
It is also worth noting that in AnchoredCLIP, $\rho(E)$ and $n_{e}(\rv)$ are never explicitly aligned (since the pairwise losses are computed only on combinations that include $C$; see \hyperref[sec:overview_methods]{Methods}); ``DOS--charge density'' nevertheless achieves reasonably good retrieval performance. 

Next, we explore how MultiMat can be used to discover materials when the desired target is not contained within the search space, by considering all DOS samples in the test set to be targets and all crystals in the train set to be potential candidates. 
Since a material with the exact target property does not exist in the search space, effectiveness is measured by how well the property of a selected material resembles the desired target. 
\Cref{fig:inverse_design}b shows the error between the DOS corresponding to the best candidate material and the desired target for this task, averaged over all targets. 
When picking the best crystal candidate out of the \( n \) closest neighbors in the shared latent space, we see that the normalized mean absolute error (MAE) between the target $\rho(E)$ and the $\rho(E)$ corresponding to the best crystal structure decreases as more neighbors are considered, as expected. 
There are diminishing improvements in normalized MAE beyond 5 neighbours, suggesting that a consideration of approximately 5 nearest neighbours would give a reasonably good candidate for the desired property. 
We expect this general trend to roughly hold when scaling to larger databases with the aim of discovering new materials with suitable properties.
Finally, \cref{fig:inverse_design}c provides a visualization of two examples from our material discovery pipeline, showing a relatively good fit between the selected material and the target $\rho(E)$.

The alignment between modalities MultiMat optimizes for ensures that a close match between $C$ and $\rho(E)$ embeddings in the multimodal space signifies similarity in the physical space between the candidate material corresponding to the $C$ embeddings and the material corresponding to the target $\rho(E)$. 
This proposed material discovery approach leverages the extensive scale of $C$ databases, which typically exceeds the number of entries for other modalities by at least an order of magnitude, and thus allows one to identify existing materials that would have a $\rho(E)$ very similar to the target, had it been computed. 
This constitutes an accelerated form of material design, which only uses inference through the neural network encoders followed by a nearest neighbor search to find materials likely to exhibit certain desired properties. This material discovery approach is enabled by the alignment between encoders that MultiMat optimizes. In contrast ``forward-only'' approaches to material design are based on an encoder-decoder structure (\eg where $C$ is first encoded to a latent space and then decoded to predict $\rho(E)$). A potential benefit of our latent-based similarity approach lies in the fact that searching for candidates in a low-dimensional latent space compared to the physical space is likely easier for high-dimensional properties such as $\rho(E)$. 
A related work focusing on $\rho(E)$ was introduced in Ref.~\citenum{bang2024inverse}; however, it differs from ours by only working for binary composition materials and focusing on material design through chemical composition for a fixed atomic structure, thus neglecting the structural information of materials.
Note that while our results focus on $\rho(E)$, the approach is applicable to other modalities, provided that the respective encoders are trained with MultiMat.

\begin{figure*}[htb]
	\centering
	\includegraphics[scale=.875]{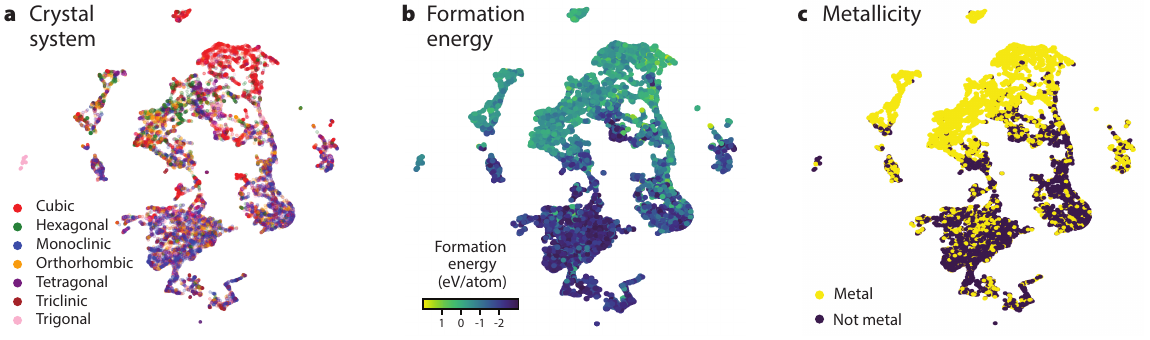}
	\caption{
        \textbf{Interpretability of crystal embeddings.}
        \textbf{a},~Crystal embeddings after dimensionality reduction by UMAP are shown, with each embedding color-coded by one of the seven crystal systems. Some clustering based on the crystal system can be observed. 
        \textbf{b},~Visualization of these dimensionality-reduced embeddings after color-coding according to each material's formation energy. 
        \textbf{c},~Visualization of these dimensionality-reduced embeddings after color-coding based on whether each material is a metal or not.
        }
    \label{fig:interpreting_embeddings}
\end{figure*}

\subsection{Interpretability of MultiMat Features}

Finally, we explore the interpretability of the MultiMat latent space.
Specifically, we use Uniform Manifold Approximation and Projection (UMAP) to transform the high-dimensional learned features from the crystal encoder into a more visualizable two-dimensional space~\citep{mcinnes2020umap}, as shown in \Cref{fig:interpreting_embeddings}.
Note that for the results in this section, we used AnchoredCLIP trained with $\{C, \rho(E), n_{e}(\rv)\}$.
We see that the 2D features reveal that materials with similar properties tend to be close together, and thus the features learned from MultiMat can be easily interpreted in a physically meaningful way.

In \cref{fig:interpreting_embeddings}a, we color-code each embedding by one of the seven possible crystal systems---cubic, hexagonal, monoclinic, orthorhombic, tetragonal, triclinic, and trigonal.
Each crystal system are collections of space groups that are typically similar to each other, thus demonstrating clustering by the spatial structure of the material.
There is some broad color-based clustering, such as cubic crystals (red) concentrating near the top and monoclinic crystals (blue) concentrating near the bottom of the 2D plot.
Additionally, some of the smaller clusters of points tend to be the same color, such as the pink cluster on the left representing a trigonal lattice.

Furthermore, we explore how the MultiMat embeddings cluster based on formation energy (a continuous property) and whether a crystal is a metal (a discrete property).
Specifically, in \cref{fig:interpreting_embeddings}b--c, we color code the dimensionality-reduced embeddings based on the value of the respective property for each material.
Although the pre-trained model has never seen labels indicating a material's formation energy or whether a material is a metal, materials with similar (different) properties are still close together (far apart) in the 2D space. 
This suggests that the model is not merely learning random abstract features or memorizing data; it is learning features that capture information about materials' physical properties. 
In future work, insights derived from these features may be used to guide the search and discovery of materials with particular optical or electronic properties without the need for costly beyond-DFT methods~\cite{knosgaard2022representing, deslippe2012berkeleygw}. 

\section{Discussion}
The incorporation of additional modalities into MultiMat improves its predictive performance. 
In particular, there is a big jump in performance between one and two modalities (\ie between the baseline and MultiMat with two modalities) and a smaller jump between two and three modalities at which point the performance improvements due to incorporating additional modalities saturates. 
This also points to promising future research opportunities in using more than two modalities for multimodal learning in domains outside of materials.

The material property prediction tasks considered in this work have less available data than what is used for the multimodal pre-training phase.
This significant difference in dataset size underscores the robust representation MultiMat develops during its pre-training phase, which likely contributes to its strong performance in crystal property prediction tasks, even with relatively limited fine-tuning data.
Small datasets are of particular interest in materials science, since many open questions in the field concern specific classes of materials with few known data points~\cite{zhang2018strategy,xu2023small, weng2020simple}. 
MultiMat could potentially alleviate some problems of traditional data-driven ML methods for materials that typically require large quantities of data.

The methodological innovations introduced in this work may also have applications beyond the domain of materials science. 
The field of multimodal learning has so far been predominantly centered on integrating just two modalities, stemming partly from the limited methodologies capable of scaling to more than two modalities~\cite{radford2021learning, li2021align, pramanick2023volta}.
Prior research in the area has largely focused on working with image--text pairs scraped from the web, thereby reducing the need for multimodal methods that go beyond two modalities~\cite{kim2021vilt, wang2022simvlm}. 
In this work, we made use of CLIP but also introduced novel extensions for multimodal pre-training that were specifically tailored to handle more than two modalities.
Furthermore, we extend our contributions by detailing two additional pre-training methods in the Supplementary Information, employing simultaneous rather than pairwise alignment of modalities, and performing on par.

An advantage of our screening approach to material discovery is the ability to constrain the search space to materials that are known to be stable.
This is a practical strategy since crystalline structures for stable materials are abundant compared to other modalities that are collected via computational methods (\eg charge density or DOS).
Our material discovery approach provides a rapid solution to material discovery and mitigates the large computational costs otherwise required in traditional simulation and experimental procedures when searching over these crystal databases.
The screening approach can also be extended to incorporate multiple modalities simultaneously---this multimodality conditioning could \eg be leveraged to identify materials with desirable properties of multiple modalities simultaneously (\eg the DOS and charge density).
Future work could explore building generative models from MultiMat's latent space to harness its effective learned representations. 
Moreover, we expect the results to further improve if the search for candidate materials is extended to larger databases of stable materials. Consequently, this represents an interesting direction for future work in materials discovery and design.
E.g., the recent GNoME database~\cite{merchant2023scaling}, consisting of 2.2 million materials predicted to be stable by ML, is particularly well-suited for this purpose. 
Other suitable databases include the Crystallography Open Database~\cite{Grazulis2012} and the Inorganic Crystal Structure Database~\cite{hellenbrandt2004inorganic}, with roughly \num{500000} and \num{280000} entries, respectively.

\section{Methods}\label{sec:methods}

\subsection{Encoder Architectures}

Here we describe the encoder architectures used for the various modalities. 

\paragraph{Crystal structure encoder}
For the $C$ encoder, we adopted the PotNet architecture~\cite{lin2023efficient}, the state-of-the-art for predicting properties of crystalline materials. PotNet represents the crystal structure data as a graph, where the nodes are atoms and the edges are interatomic potentials. In contrast to other methods, PotNet accounts for the complete set of interatomic potentials, enabling it to learn powerful representations of crystal structures. 

\paragraph{Density of states (DOS) encoder}
The data for each material consists of a list of energies $E$ and the corresponding DOS $\rho(E)$.
We utilized a Transformer architecture to encode $\rho(E)$~\citep{vaswani2023attention}. 
Because the energies $E$ for which $\rho(E)$ is measured can vary between different samples in the data, we removed the positional encoding traditionally used in Transformers and instead introduced a learnable embedding layer for the energies. 
Specifically, we separately embedded the $\rho(E)$ values and their corresponding energies $E$, followed by concatenating these embeddings along the embedding dimension (thus doubling the effective embedding dimension). Subsequently, a linear layer was employed to mix the embeddings for each token.
This was then followed by another layer, which down-sampled the embeddings for each token back to the original embedding dimension (\ie the embedding dimension is halved). 
This adaptation allows the $\rho(E)$ encoder to adeptly handle $\rho(E)$ samples with variable energy ranges since it accounts for continuous inputs (instead of discrete) and has a notion of where a particular $\rho(E)$ lies along the energy axis. 

\paragraph{Charge density encoder}
The $n_{e}(\rv)$ is represented as a three-dimensional tensor corresponding to the voxelized $n_{e}(\rv)$ (\ie a three-dimensional array of real numbers corresponding to the charge density per unit volume).
For the $n_{e}(\rv)$ encoder, we utilized a 3D ResNext architecture~\citep{xie2017aggregated} which, due to its 3D convolutions,  can capture spatial patterns in all three dimensions of the three-dimensional $n_{e}(\rv)$ tensor.

\paragraph{Text encoder}
The textual descriptions of the crystal structure are machine-generated by Robocrystallographer~\cite{robocrystallographer} and are available in the Materials Project database, similar to that used in~\cite{rubungo2023llm}. 
Each crystal is described in a paragraph containing natural language and chemical symbols. 
For better contextual understanding (in contrast to regular text models pre-trained on the Internet), we use MatBERT~\cite{matbert}, which has been pre-trained on a large corpus of material science literature, to generate embeddings for each textual description. 
MatBERT has a context window of 512 tokens; thus we truncate samples with more tokens to fit within the context window. 
Note that approximately 66\% of the dataset has less than or equal to 512 tokens and does not require truncation. 
As with most classification applications of BERT~\cite{devlin2019bert} models, the model outputs a ``CLS'' token that is typically used for downstream tasks. In this work, we use the embedding of the ``CLS'' token output as the embedding. 
Its embedding dimension is 768; to align it with the embeddings from the other encoders, we use a two-layer trainable MLP to project it down to a dimension of 128.
Note that during MultiMat pre-training, the MatBERT model is frozen (weights are not trained) and the only trainable parameters are those of the projection MLP.

\subsection{Multimodal Pre-training Methods}

CLIP~\cite{radford2021learning} is a pre-training method that makes use of image--caption pairs from the web to build effective visual representations of input text. 
CLIP makes use of a contrastive loss function to pull matching image--caption pairs (pairs where the caption corresponds to the image) closer in the embedding space whilst pushing non-matching pairs (pairs where the caption does not correspond to the image) further apart, thereby aligning the image encoder with the text encoder. Here, alignment refers to the degree to which embeddings of a matching pair of modalities are similar in the embedding space. 
This alignment results in effective visual representations that can be used for a variety of tasks~\cite{chen2020simple, radford2021learning}. 

Here, we describe the methods for multimodal pre-training that we used; first, we explain how we adapted CLIP~\cite{radford2021learning} to the domain of crystalline materials. After that, we describe our novel methods to handle multimodal pre-training with more than two modalities. In particular, we show how CLIP, which is limited to pre-training with two modalities, can be generalized to handle more than two modalities~\cite{radford2021learning}.

In the original CLIP method, we have two modalities \( \mathcal{A} \) and \( \mathcal{B} \), and their corresponding samples \( \mathbf{A}_i \) and \( \mathbf{B}_i \) for a batch of \( N \) samples (where $i$ is the index over the batch).
After the samples are encoded using the modality-specific encoders \( f_{\mathcal{A}} \) and \( f_{\mathcal{B}} \), the embeddings are given by \( \mathbf{a}_i = f_{\mathcal{A}} (\mathbf{A}_i) \) and \( \mathbf{b}_i = f_{\mathcal{B}} (\mathbf{B}_i) \). The CLIP objective connecting \( \mathcal{A} \) and \( \mathcal{B} \) is then given by
\begin{subequations}
\begin{equation}
   \ell(\mathcal{A},\mathcal{B}) 
   =
   -\sum_{i=1}^{N} \log \frac{\e^{\mathop{\mathrm{sim}}(\mathbf{a}_i, \mathbf{b}_i) / \tau}}{\sum_{j=1}^{N} \e^{\mathop{\mathrm{sim}}(\mathbf{a}_i, \mathbf{b}_j) / \tau}},
\label{eq:CLIP_single}
\end{equation}
where \(\mathop{\mathrm{sim}}(\cdot, \cdot)\) is the cosine similarity metric and \( \tau \) is the temperature parameter.
In practice, the symmetric loss
\begin{equation}
    L(\mathcal{A}, \mathcal{B}) 
    = 
    \tfrac{1}{2}
    \big[
    \ell(\mathcal{A},\mathcal{B}) + \ell(\mathcal{B},\mathcal{A})
    \big],
\label{eq:CLIP}
\end{equation}
\end{subequations}
is used. CLIP was originally introduced in the context of image--caption pairs, with \( \mathcal{A} \) representing an image modality and \( \mathcal{B} \) a text modality.

\paragraph{CLIP Adapted to Materials Science}
The most straightforward approach to multimodal pre-training in materials science is the direct adaptation of two-modality CLIP to materials-specific modalities. 
In particular, $C$ can be seen as analogous to an image and the $\rho(E)$, $n_{e}(\rv)$ or $T$ can be seen as analogous to the caption of an image in the original formulation of CLIP.
This allows us to explore three distinct options for multimodal pre-training using CLIP in materials science by making use of $C$ and $\rho(E)$, by making use of the $C$ and $n_{e}(\rv)$ or by making use of $C$ and $T$. Specifically, the loss functions are
\begin{subequations}
\begin{alignat}{2}
    &L(C, \rho),  \qquad\qquad&&\text{(crystal--DOS)}
    \\
    &L(C, n_{e}), \qquad\qquad &&\text{(crystal--charge density)}
    \\
    &L(C, T), \qquad\qquad &&\text{(crystal--text)}
\end{alignat}
\end{subequations}
where the loss function \( L \) is given by \cref{eq:CLIP}.

\paragraph{AllPairsCLIP}
Apart from a direct adaptation of CLIP to the materials science context, we also introduce two methods that extend the CLIP objective to accommodate and align an arbitrary number of modalities.
The first of these, AllPairsCLIP, generalizes the CLIP objective to more than two modalities by aggregating the CLIP losses between all combinations of two modalities.
Specifically, to incorporate all four modalities, $C$, $\rho(E)$, $n_{e}(\rv)$, and $T$, the AllPairsCLIP objective is computed as
\begin{equation}\label{eq:allpairsclip}
\begin{split}
L_{\text{AllPairsCLIP}}
=
\tfrac{1}{6}\big[ &
 L({C}, {\rho}) + L({C}, {n_{e}}) + L({C}, {T}) + {} \\
& L(\rho, n_{e}) 
+ L(\rho, {T})
+ L(n_{e}, {T})\big]
\end{split}
\end{equation}
where each term in the total loss is the individual CLIP for two modalities given by \cref{eq:CLIP}. 
A computational challenge arises from the combinatorial nature of pairwise alignments: for \( n \) modalities, the number of pairwise alignments or terms in the loss function scales as $(n^2 - n)/2$. 
This scaling is increasingly burdensome as $n$ grows. 

\paragraph{AnchoredCLIP}
To address the computational drawback posed by the AllPairsCLIP method, we propose an alternative approach, also based on CLIP, which we call AnchoredCLIP.
This method introduces the concept of an \emph{anchor modality}, a core modality, rich in information, with which every other modality shares an information-overlap with. Contrary to aligning every possible pair of modalities as in AllPairsCLIP, AnchoredCLIP only aligns pairs consisting of the anchor modality and each of the other modalities. This approach significantly reduces the number of modality-pairs being aligned, \ie terms in the loss function. Specifically, for \( n \)
modalities, the number of pairs aligned is reduced to \( n -1 \). 
In the context of materials science, when considering $C$, $\rho(E)$, $n_{e}(\rv)$, and $T$, we choose as anchor modality $C$ since it constitutes a natural representation for crystalline materials that are commonly used for downstream tasks.
The AnchoredCLIP objective for these modalities is then
\begin{equation}\label{eq:anchoredclip}
    L_{\text{AnchoredCLIP}}
    =
    \tfrac{1}{3} \big[ L({C}, \rho) + L({C}, n_{e})+ L({C}, {T}) \big],
\end{equation}
where both terms in the total loss objective are again given by the CLIP loss function in \cref{eq:CLIP}.

\paragraph{Batch masking}
When using three or more modalities via AllPairsCLIP and AnchoredCLIP, some samples may not have data entries for all the modalities---\eg some samples in the batch may have data entries for all modalities $C$, $\rho(E)$, $n_{e}(\rv)$, $T$, while some samples may have missing entries of $\rho(E)$ or $n_{e}(\rv)$ (in the Materials Project database, $C$ and $T$ exist for all the entries).
Out of \num{154718} materials in the Materials Project, there are \num{121915} with entries for $n_{e}(\rv)$, \num{89071} entries with $\rho(E)$ and \num{78461} entries with both $n_{e}(\rv)$ and $\rho(E)$. Note that $T$ exists for all $C$.
To take care of this during MultiMat pre-training, for each sampled batch of size $B$, we create a separate binary mask of dimension $B$ for each pair of modalities to indicate the existence of their data entries for each sample in the batch.
This binary mask is then used to screen and select all existing samples within the batch to compute each pair-wise loss while setting the loss terms of the missing entries to zero, thus batch-wise training can be performed as per normal. 

\subsection{Material Discovery via Latent Space Similarity and Interpretability of Embeddings}
Here, we elaborate on the experimental procedures undertaken for the results pertaining to material discovery and the interpretability analysis of embeddings following multimodal pre-training. For the retrieval and material discovery experiments illustrated in \cref{fig:inverse_design}, we utilized encoders that were pre-trained using AnchoredCLIP on three modalities of $C$, $\rho(E)$ and $n_{e}(\rv)$. 
We split the pre-training dataset into train--test subsets in an 80:20 ratio (resulting in approximately \num{62000} and \num{16000} train and test materials respectively).  
MultiMat pre-training was performed on the training set and the retrieval accuracy shown in \cref{fig:inverse_design}a was computed on the test set (\ie consisting of samples not in the train set which was used for the multimodal pre-training). Regarding the experiments showcased in \cref{fig:inverse_design}b-c, the target $\rho(E)$ came from the test set, again ensuring these were not part of the pre-training dataset.
We then treated all materials in the train set as potential candidate materials, aiming to identify the materials being the closest neighbors for each target $\rho(E)$.

For the quantitative evaluation of the material discovery strategy shown in \cref{fig:inverse_design}b, we compute the MAE between the target and nearest-neighbor $\rho(E)$ in the energy range from $-5\,\text{eV}$ to $+5\,\text{eV}$, using linear interpolation to map the target and nearest-neighbor $\rho(E)$ onto the same equispaced energy grid.
We restrict our focus to this limited range because it 
(i)~helps to account for the varying energy ranges of different materials in the Materials Project data, obviating a need for extrapolation, and
(ii)~covers the energy range of primary physical interest, since most electrical and optical properties are influenced mainly by electrons near the Fermi level~\cite{mahan2000many, grosso2013solid, kong2022density}. 
Additionally, the MAE between the target and nearest neighbor $\rho(E)$ was normalized by the area of the target $\rho(E)$ in the $-5\,\text{eV}$ to $+5\,\text{eV}$ range. 
This normalization ensures a more equitable comparison across different targets--nearest-neighbor pairs. 
Mathematically, we define the normalized MAE in the energy range from $-5\,\text{eV}$ to $+5\,\text{eV}$ by
\begin{equation} 
    \text{nMAE}
    =
    \frac{\int_{-5\,\text{eV}}^{5\,\text{eV}} |\rho_{\text{target}}(E) - \rho_{\text{nearest neighbour}}(E)| \, \mathrm{d}E}{\int_{-5\,\text{eV}}^{5\,\text{eV}} \rho_{\text{target}}(E) \, \mathrm{d}E}.
\end{equation}
 
Note that this metric, despite its relative character, may still exhibit large values (\eg exceeding unity), even for a slight misalignment of the resonance energies because the DOS frequently is a sharply peaked quantity.

For the interpretability results presented in \cref{fig:interpreting_embeddings}, we made use of materials from the same test set that was used for the retrieval and material discovery results discussed above. The embeddings of the approximately \num{16000} crystal structures in the test set were transformed into a two-dimensional space through UMAP dimensionality reduction~\cite{mcinnes2020umap}. In \cref{fig:interpreting_embeddings}b, a few of these materials were identified as outliers in terms of their formation energy and thus removed. This was done to make the color gradient easier to interpret. 

\subsection{Data}
We constructed a multimodal dataset for materials science using data from the Materials Project~\cite{10.1063/1.4812323}, a well-established open-source initiative.
This dataset included crystal structures, density of states, charge densities, and textual descriptions; those four modalities were used for multimodal pre-training. In addition to those modalities, we also made use of the bulk modulus, shear modulus, and elastic tensor data for material property prediction performance evaluation of MultiMat (after fine-tuning on those tasks) as well as for establishing non-pre-trained baselines. We also used Materials Project data for the interpretability results where we color-coded by crystal system, formation energy, and whether the material is a metal.

Despite its comprehensiveness, the Materials Project has known data quality limitations for certain material properties, \eg for the band gaps.
Specifically, the RMSE between the Materials Project band gaps (computed using DFT) and their experimentally observed counterparts is \(1.05\) eV, potentially affecting the efficacy and reliability of models trained on band gaps from the Materials Project~\cite{10.1063/1.4812323}.
To address this, we utilized the HSE gaps in the SNUMAT semiconductor database~\cite{article_snumat}, which offers more accurate band gap values (RMSE of $0.36$ eV relative to experimentally determined band gaps) due to using a more accurate DFT functional.
We used the version of this database where materials with a computed gap of \SI{0}{eV} were filtered out; note that even this filtered version of this database does contain some large gap insulators.
This SNUMAT semiconductor database contains around \num{10000} materials without any multimodal information.
We used it to fine-tune and evaluate models pre-trained with multimodal data from the Materials Project and also to establish baselines for models without any multimodal pre-training.
Note that some previous works~\cite{wang2021accurate,choubisa2023interpretable} have explored using ML to predict band gaps of the SNUMAT database.

\subsection{Implementation Details and Settings for Training and Evaluation}
\paragraph{MultiMat pre-training} 
We use the PotNet architecture for the $C$ encoder, a Transformer-based architecture for the $\rho(E)$ encoder, a 3D ResNeXt architecture for the $n_{e}(\rv)$ encoder, and MatBERT~\cite{matbert} (together with a two-layer MLP) for the $T$ encoder.
Each encoder produces an embedding with dimension $d=128$.
We use the AdamW optimizer~\citep{loshchilov2018decoupled} for training, with a cosine-decay learning rate schedule and a linear warm-up schedule of 10 epochs. 
The peak learning rate is fixed at $10^{-4}$ and weight-decay is fixed at \num{5e-4}. We use a batch size of 360 across all pre-training experiments and perform pre-training for a total of 500 epochs. 

\paragraph{Fine-tuning for prediction tasks}
After pre-training, the $C$ encoder is transferred, and a linear head is randomly initialized. The model was then fine-tuned for various material property prediction tasks.
We use the AdamW optimizer with a cosine-decay learning rate schedule and linear warm-up with \num{10} epochs.
We use a batch size of \num{120}, no weight-decay, and the peak learning rate was swept over $\{10^{-3}, 10^{-4}, 10^{-5}\}$.
From the downstream data entries available for the specific prediction task, we create a train, validation, and test split in the ratio of $60\,{:}\,20\,{:}\,20$.
The pre-trained $C$ encoder was fine-tuned on the training set and early stopping was performed based on the lowest validation error on the validation set.
The best checkpoint (\ie with the lowest validation loss) was then used to evaluate on the test set.
Error bars were created by taking the standard deviation from three different experiments with different seeds.

\paragraph{Material discovery via latent space similarity}
For the results in \cref{fig:inverse_design}, we used a slightly smaller batch size of 100 for MultiMat pre-training as we observed that this resulted in slightly better performance.

\subsection{Data availability}
This article uses data that are all open-sourced and publicly available. Specifically, the data is downloaded from the Materials Project (\url{https://next-gen.materialsproject.org/}) and SNUMAT (\url{https://www.snumat.com/}) databases. 

\subsection{Code availability}
MultiMat was developed using the PyTorch framework. All source codes used for training and for generating all the results are publicly available at \url{https://github.com/vmoro1/multimat}.

%----------------------------
%----- ACKNOWLEDGEMENTS -----
%----------------------------

\section*{Acknowledgements}
We thank Sean Mann, Michael Huang, Donato Jimenez Beneto, Di Luo, Owen Dugan, Li Jing, Jasper Snoek, and Jamie Smith for fruitful discussions.
This research was sponsored in part by the United States Air Force Research Laboratory and the Department of the Air Force Artificial Intelligence Accelerator and was accomplished under Cooperative Agreement Number FA8750-19-2-1000. The views and conclusions contained in this document are those of the authors and should not be interpreted as representing the official policies, either expressed or implied, of the Department of the Air Force or the U.S.\ Government. The U.S.\ Government is authorized to reproduce and distribute reprints for Government purposes notwithstanding any copyright notation herein.
This material is also based upon work sponsored in part by the U.S.\ Army DEVCOM ARL Army Research Office through the MIT Institute for Soldier Nanotechnologies under Cooperative Agreement number W911NF-23-2-0121, and in part by the Air Force Office of Scientific Research under the award number FA9550-21-1-0317.
T.C.\ acknowledges the support of a research grant (project no.~42106) from Villum Fonden. 
C.L.\ received support from DSO National Laboratories of Singapore. 
A.M.\ received support from the National Science Foundation Graduate Research Fellowship under Grant No.~1745302.
P.Y.L.~gratefully acknowledges the support of the Eric and Wendy Schmidt AI in Science Postdoctoral Fellowship, a Schmidt Sciences program.

%----------------------
%----- REFERENCES -----
%----------------------
\FloatBarrier
\bibliographystyle{apsrev4-2-longbib}
\bibliography{main}

%----------------------------
%----- APPENDIX -----
%----------------------------
\cleardoublepage

\end{document}